\documentclass{article}
\usepackage{spconf}
\usepackage[T1]{fontenc}
\usepackage{amsmath,amssymb,graphicx,booktabs,cite,url,microtype}
\usepackage{multirow}
\title{QCVS: Question-Contrastive Visual Steering for Multimodal Reasoning}
\name{Tong Fu$^{1,2}$}
\address{%
  $^{1}$Guangxi Key Laboratory of Digital Infrastructure \\
  $^{2}$Xi'an Jiaotong University
}
\begin{document}
\ninept
\maketitle

\begin{abstract}
Inference-time steering in vision-language models often entangles query-relevant reasoning dynamics with low-level perceptual noise, causing collateral degradation on correct predictions. We propose Question-Contrastive Visual Steering (QCVS) to isolate task-specific intervention directions via double-differential activation probing. By contrasting spatial layout perturbations under the target query against a generic description reference, QCVS cancels task-agnostic visual artifacts without ground-truth supervision. A dual-threshold gating mechanism then evaluates relative interaction strength and signal amplitude, selectively applying a single bounded hidden-state update only when necessary. Empirical evaluations across multiple scales of Qwen3.5 (2B to 9B) and DeepSeek-VL2 on DynaMath and We-Math demonstrate that QCVS achieves state-of-the-art accuracy gains under sparse coverage ($\approx 22\%\text{--}30\%$). Crucially, QCVS yields repair rates up to $13.98\%$ while suppressing collateral damage down to $0.41\%$, proving high intervention precision and computational efficiency.
\end{abstract}


\section{Introduction}
\label{sec:introduction}
Visual reasoning requires dynamically isolating question-relevant visual evidence rather than merely recognizing generic scene content~\cite{clevr,gqa,spark}. In complex multimodal mathematical reasoning~\cite{vskip,zhang2026not,mathvista,mathvision}, distinct queries demand contrasting visual features from the same image to support multi-step symbolic inference. However, most existing inference-time intervention methods~\cite{vcd,vista} evaluate visual perturbations solely under the task prompt. This conflates query-guided reasoning signals with task-agnostic visual geometry and perturbation sensitivity, inadvertently amplifying low-level perceptual noise and causing collateral performance degradation on initially correct predictions. Furthermore, deciding when to intervene~\cite{octopus,viti,caac} remains fundamentally distinct from constructing a precise steering vector.

To address this critical limitation, we introduce \emph{Question-Contrastive Visual Steering} (QCVS) to isolate task-specific steering directions without model retraining via double-differential probing. QCVS applies identical spatial perturbations under both the target task question and a fixed description reference prompt. By subtracting the reference activation response from the task response, QCVS effectively cancels shared spurious background visual artifacts, isolating an input-specific interaction direction without requiring any ground-truth labels or output confidence statistics.

\begin{figure*}[t]
\centering
\includegraphics[width=\textwidth]{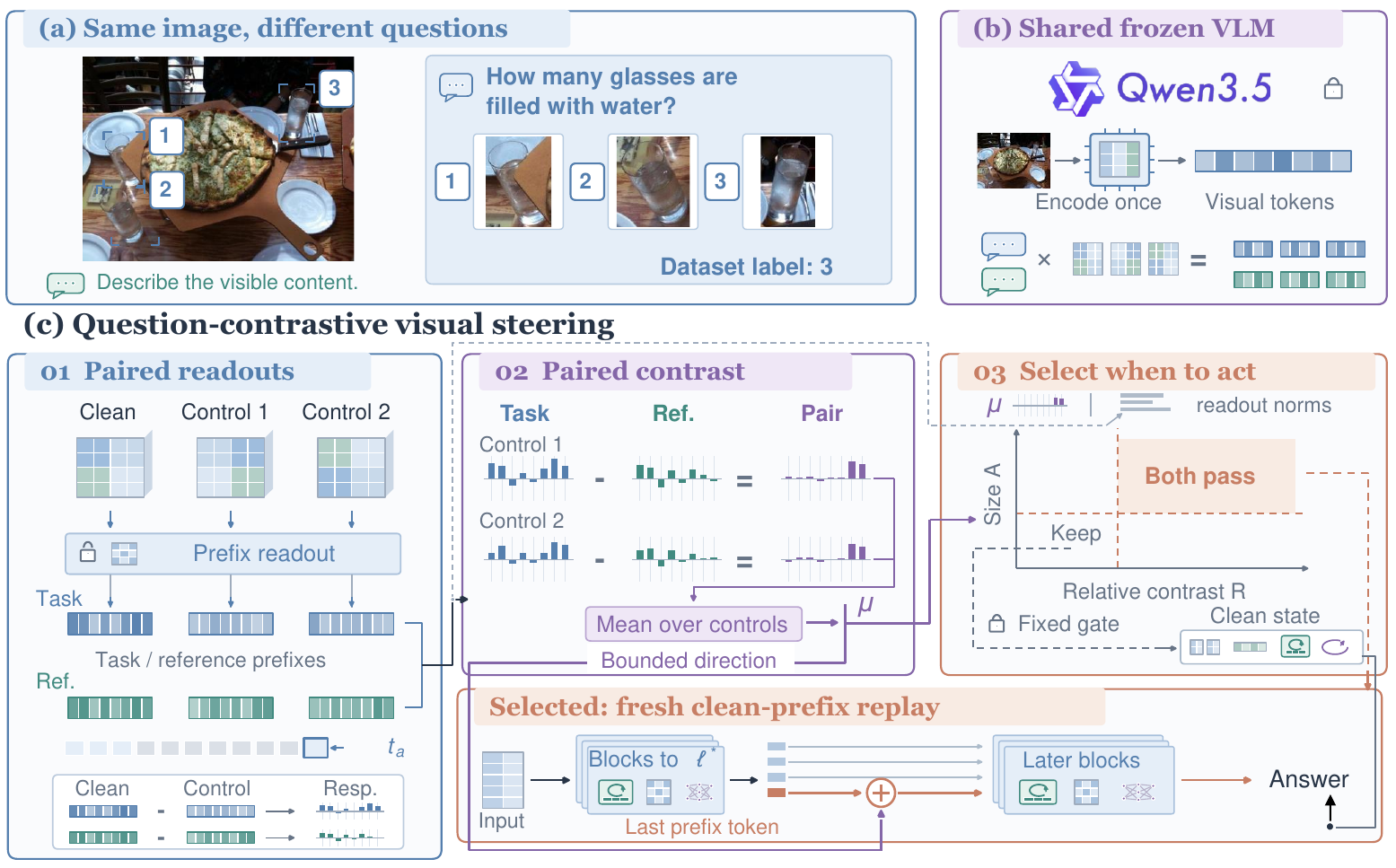}
\caption{Motivation and computation of QCVS. (a) Describing a scene and counting water-filled glasses require different evidence. The BLINK~\cite{blink} example has dataset label 3; the crops are human-selected illustrations, not model outputs. (b) One visual encoding feeds a frozen Qwen3.5 model: two questions and three visual conditions give six prefix readouts. (c) Paired clean--control differences at each final assistant-prefix token $t_a$ yield the interaction $\mu$. Its direction defines a bounded edit; $R$ and $A$ determine selection. A selected input receives one update at $(\ell^*,t_a)$ in a fresh clean-prefix replay; otherwise, decoding uses the retained clean hybrid state.}
\label{fig:overview}
\end{figure*}

As illustrated in Figure~\ref{fig:overview}, QCVS executes contrastive direction through a selective, single-site intervention mechanism. Two hidden-state magnitude ratios evaluate relative interaction strength and signal amplitude against calibration-derived thresholds to govern intervention gating. When triggered, a single bounded update is injected into the target state during a fresh prefix replay pass before generation, leaving unselected inputs on their original decoding paths. With all model parameters remaining frozen, this strategy ensures precise activation alignment while strictly guarding against unhelpful edits.

Our primary contribution encompasses both the paired direction construction and its integration into a selective single-site inference procedure. We establish a double-differential probing paradigm that eliminates spurious task-agnostic visual noise, couple it with a dual-threshold gating mechanism that restricts state edits to beneficial cases, and formulate a rigorous, fine-grained evaluation protocol accounting for accuracy, repair, damage, and coverage. This systematic framework explicitly separates the intrinsic utility of reference subtraction from the effect of sparse intervention coverage.

\section{Related work}
Inference-time interventions for vision--language models primarily operate on output logit distributions~\cite{m3id,avisc,opera} or internal hidden activations~\cite{vti,pti,caa}. Visual contrastive decoding compares predictions under original and distorted images to mitigate object hallucinations~\cite{vcd}. VISTA combines visual activation steering with early-layer token-logit augmentation to guide text generation~\cite{vista}. Octopus dynamically learns to select among contrastive decoding strategies and a non-intervention option at each generation step~\cite{octopus}. For text-based reasoning, SPARK uses susceptibility profiling to identify latent states and construct steering directions~\cite{spark}. Unlike existing methods, QCVS focuses on the principled construction of question-contrastive visual directions: it compares activation responses to matched visual perturbations under a task question and a generic-description reference.

\section{Method}
\label{sec:method}

\subsection{Matched Visual Controls}

A single forward pass through the vision encoder and projection layer yields the visual token grid $Z \in \mathbb{R}^{N \times d}$, where $N$ denotes the total number of visual tokens and $d$ represents the hidden feature dimension. We instantiate $K=2$ matched controls $Z_k = P_k Z$ by permuting non-overlapping $2 \times 2$ token blocks within the largest top-left even subgrid, while holding boundary strips and internal block structures invariant. Sample-keyed permutation matrices $P_k$ are shared between the target task question $q$ and a fixed reference prompt $q_r$ defined as \emph{Describe the visible content}. Inputs containing fewer than two movable visual blocks automatically bypass steering. This spatial permutation preserves the multiset of visual token vectors, token counts, position slots, and sequence masks, establishing a matched feature perturbation rather than a semantic counterfactual. Let $h_\ell(Z,p) \in \mathbb{R}^d$ denote the post-block hidden representation at the final assistant-prefix position $t_a$ under question $p$, located independently within each architecture-specific chat template. Evaluating both question prompts under $Z, Z_1, Z_2$ requires six prefix evaluations without generating auto-regressive answer continuations.

\subsection{Question-Contrastive Direction Derivation}

To isolate task-relevant multimodal reasoning dynamics from low-level visual processing artifacts, we compute differential response vectors that capture internal activation shifts under structured spatial layout perturbations across both question conditions:
\begin{equation}
d_{\ell k}^{q} = h_\ell(Z,q) - h_\ell(Z_k,q), \quad d_{\ell k}^{r} = h_\ell(Z,q_r) - h_\ell(Z_k,q_r).
\label{eq:qcvs_responses}
\end{equation}

Here, $d_{\ell k}^{q}$ reflects joint multimodal reasoning updates and spatial layout disruptions, whereas $d_{\ell k}^{r}$ isolates task-agnostic perceptual sensitivity shifts under a fixed generic descriptive query $q_r$. To eliminate task-independent visual noise, we subtract the reference response from the task response and average across all $K$ perturbation controls:
\begin{equation}
\Delta_{\ell k} = d_{\ell k}^{q} - d_{\ell k}^{r}, \quad \mu_\ell = \frac{1}{K}\sum_{k=1}^{K}\Delta_{\ell k}.
\label{eq:pair}
\end{equation}

Reference subtraction effectively cancels shared spurious background artifacts, isolating query-guided reasoning updates without requiring ground-truth labels. Averaging over controls suppresses spatial permutation variance, unlike standard task-only steering that relies solely on the uncalibrated average $K^{-1}\sum_k d_{\ell k}^{q}$.

\begin{table*}[t]
\caption{Comparison across three Qwen3.5 model sizes on DynaMath and We-Math. Acc.\ is the percentage of input questions answered correctly. Rep.\ is the percentage of questions answered incorrectly by Base that become correct after applying a method. Dmg.\ is the percentage of questions answered correctly by Base that become incorrect after applying a method. Cov.\ is the percentage of inputs on which the method applies an intervention. Tokens is the mean number of generated tokens per input, including reasoning and the final answer. For each model on each dataset, the highest accuracy is bolded across all methods; the highest repair rate and lowest damage rate are bolded among non-Base methods. Base is excluded from the latter comparisons because it performs no intervention.}
\label{tab:main}
\centering
\setlength{\tabcolsep}{3pt}
\renewcommand{\arraystretch}{1.08}
\begin{tabular*}{\textwidth}{@{\extracolsep{\fill}}cl*{10}{c}@{}}
\toprule
\multirow{2}{*}{Model} & \multicolumn{1}{c}{\multirow{2}{*}{Method}}
& \multicolumn{5}{c}{DynaMath}
& \multicolumn{5}{c}{We-Math}\\
\cmidrule(lr){3-7}\cmidrule(l){8-12}
& & Acc.$\uparrow$ & Rep.$\uparrow$ & Dmg.$\downarrow$ & Cov. & Tokens
  & Acc.$\uparrow$ & Rep.$\uparrow$ & Dmg.$\downarrow$ & Cov. & Tokens\\
\midrule
\multirow{5}{*}{Qwen3.5-2B}
& Base & 73.59 & 0.00 & 0.00 & 0.00 & 1802
 & 66.32 & 0.00 & 0.00 & 0.00 & 2445\\
& VCD & 74.43 & 8.99 & 2.09 & 100.00 & 1885
 & 67.64 & 6.83 & 1.47 & 100.00 & 2552\\
& VISTA & 74.51 & 8.47 & 1.79 & 100.00 & 1844
 & 68.28 & 8.02 & 1.13 & 100.00 & 2498\\
& Task-only & 74.63 & 7.03 & 1.11 & 30.00 & 1852
 & 68.16 & 6.83 & 0.69 & 27.99 & 2505\\
& QCVS (Ours) & \textbf{76.11} & \textbf{12.02} & \textbf{0.90} & 30.00 & 1877
 & \textbf{69.02} & \textbf{9.22} & \textbf{0.61} & 27.99 & 2535\\
\midrule
\multirow{5}{*}{Qwen3.5-4B}
& Base & 83.29 & 0.00 & 0.00 & 0.00 & 2339
 & 75.40 & 0.00 & 0.00 & 0.00 & 3209\\
& VCD & 83.79 & 8.96 & 1.20 & 100.00 & 2465
 & 76.09 & 7.01 & 1.37 & 100.00 & 3376\\
& VISTA & 84.29 & 10.99 & 1.01 & 100.00 & 2402
 & 76.67 & 8.18 & 0.99 & 100.00 & 3293\\
& Task-only & 84.13 & 8.96 & 0.79 & 25.01 & 2402
 & 76.61 & 7.01 & 0.69 & 25.00 & 3293\\
& QCVS (Ours) & \textbf{85.21} & \textbf{13.98} & \textbf{0.50} & 25.01 & 2433
 & \textbf{77.30} & \textbf{9.35} & \textbf{0.53} & 25.00 & 3335\\
\midrule
\multirow{5}{*}{Qwen3.5-9B}
& Base & 83.59 & 0.00 & 0.00 & 0.00 & 3058
 & 75.17 & 0.00 & 0.00 & 0.00 & 4214\\
& VCD & 84.05 & 9.98 & 1.41 & 100.00 & 3231
 & 75.86 & 6.94 & 1.38 & 100.00 & 4447\\
& VISTA & 84.65 & 9.98 & 0.69 & 100.00 & 3145
 & 76.90 & 9.26 & 0.76 & 100.00 & 4330\\
& Task-only & 84.33 & 7.54 & 0.60 & 22.00 & 3134
 & 76.38 & 6.94 & \textbf{0.69} & 22.01 & 4316\\
& QCVS (Ours) & \textbf{85.35} & \textbf{12.77} & \textbf{0.41} & 22.00 & 3172
 & \textbf{77.43} & \textbf{11.41} & 0.76 & 22.01 & 4368\\
\bottomrule
\end{tabular*}
\end{table*}

\subsection{Selective Gating and Bounded Steering}

We quantify contrastive interaction relative to perturbation response and clean task state via two dimensionless ratios:
\begin{equation}
\begin{aligned}
R_\ell = \frac{\|\mu_\ell\|_2}{\frac{1}{K}\sum_{k=1}^K(\|d_{\ell k}^{q}\|_2 + \|d_{\ell k}^{r}\|_2) + \epsilon}, 
A_\ell = \frac{\|\mu_\ell\|_2}{\|h_\ell(Z,q)\|_2 + \epsilon},
\end{aligned}
\label{eq:score}
\end{equation}
where $R_\ell \in [0,1]$ measures relative interaction strength, $A_\ell$ measures signal amplitude, and $\epsilon=10^{-8}$ prevents division by zero in floating-point operations. A source-disjoint calibration set establishes activation threshold $\tau_A$ at the 25th percentile of $A_\ell$ and norm bounding interval $[s_-, s_+]$ at the 5th and 95th percentiles of $\|\mu_\ell\|_2$. Validation selects layer $\ell^*$, multiplier $\alpha$, and relative threshold $\tau_R$. For any test input, the gating decision $g(x)$ is defined as:
\begin{equation}
g(x) = \mathbf{1}\{R_{\ell^*} > \tau_R\} \cdot \mathbf{1}\{A_{\ell^*} > \tau_A\}.
\label{eq:qcvs_gate}
\end{equation}

Invalid controls force $g(x)=0$. When $g(x)=1$, the direction vector is normalized as $u = \mu_{\ell^*}/\|\mu_{\ell^*}\|_2$ and bounded via $s = \operatorname{clip}(\|\mu_{\ell^*}\|_2, s_-, s_+)$, updating the clean state during a replay pass:
\begin{equation}
h'_{\ell^*,t_a} = h_{\ell^*}(Z,q) + \alpha s u.
\label{eq:action}
\end{equation}

This update modifies a single token vector after layer $\ell^*$, allowing subsequent layers to propagate activations normally; unselected inputs ($g(x)=0$) execute standard decoding.

\subsection{Execution and Computational Complexity}

The clean task prefix is evaluated last during probing to retain its states, while auxiliary prefixes run without caching. When $g(x)=1$, generation proceeds from a fresh replay pass, restoring key-value caches, recurrent states, and metadata. We approximate total computational complexity $C(x)$ as:
\begin{equation}
C(x) \simeq C_V + (K+1)(C_q + C_r) + g(x)C_q + C_G,
\label{eq:cost}
\end{equation}
where $C_V, C_G$ denote visual encoding and generation costs, while $C_q, C_r$ represent prefix processing. Assuming equal prefix costs, probing and intervention overhead translates to $6+c$ prefix equivalents on average, where $c = \mathbb{E}[g(x)]$ denotes empirical coverage.

\section{Experiments and Analysis}
\subsection{Experimental Setup}
We evaluate our proposed approach on two challenging multimodal mathematical reasoning benchmarks, DynaMath~\cite{dynamath} and We-Math~\cite{wemath}. To investigate model scaling behavior, primary experiments span three parameter sizes of the Qwen3.5 family~\cite{qwen35} (Qwen3.5-2B, 4B, and 9B). To assess architectural transferability across backbones, we further incorporate DeepSeek-VL2 variants~\cite{deepseekvl2}, including DeepSeek-VL2-Tiny (1.0B active parameters) and DeepSeek-VL2-Small (2.8B active parameters), under model-specific calibration protocols.

We compare QCVS against the unedited Base model and representative intervention baselines, including VCD~\cite{vcd}, VISTA~\cite{vista}, and a Task-only direction baseline. Unlike full-coverage methods (VCD, VISTA) that edit all instances, gated variants apply targeted updates under controlled coverage thresholds. All baselines adopt official configurations under standard inference setups.

Evaluation spans five metrics covering error correction and operational efficiency. Overall accuracy ($\text{Acc.}$) measures the proportion of correctly solved queries. Fine-grained intervention dynamics are captured by repair rate ($\text{Rep.}$, percentage of baseline errors rectified) and damage rate ($\text{Dmg.}$, proportion of initially correct predictions degraded). Operational efficiency is quantified via coverage ($\text{Cov.}$, percentage of gated interventions) and average token count ($\text{Tokens}$, accounting for both reasoning steps and output length).

\begin{table}[t]
\centering
\caption{Cross-backbone generalization performance on the DynaMath benchmark. Experiments are conducted on DeepSeek-VL2-Tiny (1.0B active parameters) and DeepSeek-VL2-Small (2.8B active parameters) with model-specific calibration. Metrics follow Table~\ref{tab:main}. Bold marks the highest accuracy per model.}
\label{tab:transfer}
\begingroup
\setlength{\tabcolsep}{3pt}
\renewcommand{\arraystretch}{1.08}
\begin{tabular*}{\linewidth}{@{\extracolsep{\fill}}lcccc@{}}
\toprule
\multicolumn{1}{c}{Method} & Acc.$\uparrow$ & Rep.$\uparrow$ & Dmg.$\downarrow$ & Cov.\\
\midrule
\multicolumn{5}{c}{\textit{DeepSeek-VL2-Tiny (1.0B active)}}\\
Base & 45.01 & 0.00 & 0.00 & 0.00\\
Task-only & 46.19 & 3.38 & 1.51 & 30.00\\
QCVS & \textbf{47.13} & 4.86 & 1.24 & 30.00\\
\midrule
\multicolumn{5}{c}{\textit{DeepSeek-VL2-Small (2.8B active)}}\\
Base & 55.01 & 0.00 & 0.00 & 0.00\\
Task-only & 56.21 & 3.95 & 1.05 & 25.01\\
QCVS & \textbf{56.95} & 5.19 & 0.73 & 25.01\\
\bottomrule
\end{tabular*}
\endgroup
\end{table}

\subsection{Main Results}
Empirical evaluation across model scales demonstrates that targeted representation intervention systematically elevates multimodal mathematical reasoning performance while maintaining strict intervention efficiency. As detailed in Table~\ref{tab:main}, QCVS consistently achieves state-of-the-art results across all three Qwen3.5 parameter scales (2B, 4B, and 9B) on both DynaMath and We-Math benchmarks. On DynaMath, QCVS improves accuracy over the Base model by $+2.52$, $+1.92$, and $+1.76$ percentage points across the 2B, 4B, and 9B architectures, reaching a peak accuracy of $85.35\%$. Similarly, on We-Math, QCVS outperforms all competing approaches, attaining top accuracies of $69.02\%$ (2B), $77.30\%$ (4B), and $77.43\%$ (9B).

A granular analysis of error dynamics and resource consumption further illuminates the mechanisms behind these gains. Compared to full-coverage baselines such as VCD and VISTA that apply interventions across $100\%$ of inputs, QCVS operates selectively under sparse coverage ($\text{Cov.} \approx 22\%\text{--}30\%$) through its joint gating mechanism. Under this controlled regime, QCVS delivers substantial error correction, achieving $\text{Rep.}$ up to $13.98\%$ on DynaMath and $11.41\%$ on We-Math, while successfully suppressing collateral $\text{Dmg.}$ down to $0.41\%$ on the 9B model. Furthermore, inspection of output token generation shows that average token counts remain closely aligned with the unedited Base model across all sizes. This confirms that performance gains stem directly from precise internal state alignment rather than length inflation or unconstrained generation expansion.

\subsection{Sensitivity to Intervention Coverage}
\begin{figure}[t]
\centering
\includegraphics[width=\linewidth]{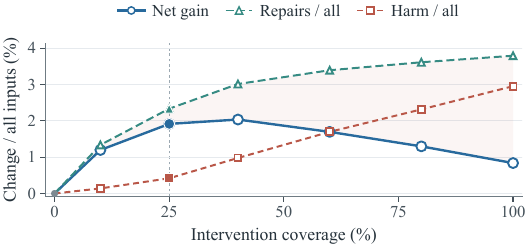}
\caption{Fixed-action coverage sweep on Qwen3.5-4B/DynaMath. Curves show repairs, harm and net change as percentages of all inputs; net change equals accuracy gain (pp).}
\label{fig:coverage}
\vspace{-2mm}
\end{figure}

Varying the gating threshold on Qwen3.5-4B exposes the trade-off between trajectory correction and collateral harm (Figure~\ref{fig:coverage}). As coverage expands toward $100\%$, raw repairs show diminishing returns while harm accumulates steadily, escalating from $0.43\%$ to $2.95\%$. Consequently, overall net accuracy gain follows a concave trajectory, peaking between $25\%$ and $40\%$ coverage. At the default threshold ($25.01\%$), QCVS achieves near-optimal net gain by capturing $2.31\%$ repairs while suppressing harm to $0.43\%$. Full $100\%$ coverage degrades net gain to $0.84\%$, demonstrating that sparse gating is essential to avoid perturbing valid reasoning pathways.

\subsection{Cross-Backbone Generalization}
We evaluate the generalizability of QCVS across distinct model architectures using DeepSeek-VL2 variants on DynaMath (Table~\ref{tab:transfer}). Across both 1.0B and 2.8B parameter scales, QCVS consistently achieves the highest overall accuracy, outperforming the Base model and Task-only baseline under equivalent coverage. These consistent gains across non-Qwen architectures indicate that the proposed vector intervention effectively corrects reasoning errors without relying on specific backbone initializations or pre-training setups.

\begin{table}[t]
\centering
\caption{Ablation of QCVS components on Qwen3.5-4B (DynaMath). $\Delta$Acc.\ is accuracy gain over Base (pp); Rep.\ and Dmg.\ are in \%. All settings use 25.01\% coverage except Ungated (100\%).}
\label{tab:ablation}
\begingroup
\setlength{\tabcolsep}{4pt}
\renewcommand{\arraystretch}{1.05}
\begin{tabular*}{\linewidth}{@{\extracolsep{\fill}}lccc@{}}
\toprule
\multicolumn{1}{c}{Variant} & $\Delta$Acc.$\uparrow$ & Rep.$\uparrow$ & Dmg.$\downarrow$\\
\midrule
\multicolumn{4}{c}{\textit{Intervention Direction}}\\
QCVS (Ours) & \textbf{+1.92} & \textbf{13.98} & \textbf{0.50}\\
Task-only direction & +0.84 & 8.96 & 0.79\\
Random direction & -0.88 & 1.91 & 1.44\\
Reversed direction & -1.86 & 1.08 & 2.44\\
\midrule
\multicolumn{4}{c}{\textit{Selection / Gating Strategy}}\\
QCVS Gate (Ours) & \textbf{+1.92} & 13.98 & \textbf{0.50}\\
Random gate & +0.22 & 5.73 & 0.89\\
$A$-only gate & +1.02 & 10.99 & 0.98\\
Ungated & +0.84 & \textbf{22.70} & 3.55\\
\bottomrule
\end{tabular*}
\endgroup
\end{table}

Analysis of error metrics shows that QCVS improves accuracy by simultaneously increasing the $\text{Rep.}$ and reducing the $\text{Dmg.}$ On DeepSeek-VL2-Tiny, QCVS increases the repair rate to $4.86\%$ while lowering the damage rate to $1.24\%$. On the larger 2.8B model, the damage rate decreases further to $0.73\%$. This pattern demonstrates that the joint gating mechanism selectively targets incorrect reasoning paths without destabilizing correctly answered inputs.

\subsection{Ablation Analysis}
Table~\ref{tab:ablation} decouples the impact of intervention direction and gating mechanism on Qwen3.5-4B. For direction design (top block), substituting our QCVS vector with a standard Task-only direction halves the accuracy gain ($\Delta\text{Acc}$ drops from $+1.92$ to $+0.84$), whereas random or reversed directions degrade baseline performance ($-0.88$ and $-1.86$), validating the semantic precision of our targeted vector. For gating strategy (bottom block), random intervention yields minimal improvement ($+0.22$), and the single-signal $A$-only gate underperforms our joint formulation ($+1.02$). Notably, while the Ungated baseline achieves high raw repairs ($\text{Rep.} = 22.70\%$), it inflicts severe side effects ($\text{Dmg.} = 3.55\%$), resulting in a degraded net accuracy gain ($+0.84$). This confirms that selective gating is essential to prune unhelpful interventions and safeguard general capability.

\section{Conclusion}
\label{sec:conclusion}
We proposed Question-Contrastive Visual Steering (QCVS), a training-free inference mechanism for vision-language models. Contrasting spatial layout perturbations against a generic reference cancels task-agnostic artifacts to isolate query-guided reasoning directions. A dual-threshold gate restricts bounded updates to beneficial instances. Across diverse architectures, QCVS achieves high repair rates while suppressing collateral damage. A current limitation is its reliance on structured spatial permutations, which may not extend to non-grid formats. Overall, double-differential probing provides an efficient paradigm for precise multimodal activation alignment.

\clearpage
\bibliographystyle{IEEEbib}
\bibliography{refs}
\end{document}